# AUTOMATIZED MARINE VESSEL MONITORING FROM SENTINEL-1 DATA USING CONVOLUTION NEURAL NETWORK

*Surya Prakash Tiwari[1], Sudhir Kumar Chaturvedi[2], Subhrangshu Adhikary[3], Saikat Banerjee[4], Sourav Basu[5]*

[1]Center for Environment & Water, Research Institute, King Fahd University of Petroleum & Minerals, Dhahran, 31261, Kingdom of Saudi Arabia
[2]Department of Aerospace Engineering, University of Petroleum and Energy Studies, Dehradun-248007, India
[3]Dr. B.C. Roy Engineering College, Durgapur-713206, West Bengal, India
[4]Department Of Mechanical Engineering, CubicX, Kolkata-700070, West Bengal, India
[5]Department Of Electrical Engineering, CubicX, Kolkata-700070, West Bengal, India

**ABSTRACT**

The advancement of multi-channel synthetic aperture radar (SAR) system is considered as an upgraded technology for surveillance activities. SAR sensors onboard provide data for coastal ocean surveillance and a view of the oceanic surface features. Vessel monitoring has earlier been performed using Constant False Alarm Rate (CFAR) algorithm which is not a smart technique as it lacks decision-making capabilities, therefore we introduce wavelet transformation-based Convolution Neural Network approach to recognize objects from SAR images during the heavy naval traffic, which corresponds to the numerous object detection. The utilized information comprises Sentinel-1 SAR-C dual-polarization data acquisitions over the western coastal zones of India and with help of the proposed technique we have obtained 95.46% detection accuracy. Utilizing this model can automate the monitoring of naval objects and recognition of foreign maritime intruders.

## 1. INTRODUCTION

The surveillance is intended to support efforts related to security, safety, environmental and sustainability aspects. Automatic Identification System (AIS) and Vessel Monitoring System (VMS) are the most reliable options among detection techniques [1]. Other detection types do not require cooperation on the vessel's side, known as the non-cooperative systems. These systems generally utilize imaging sensors for marine object detection [2]. Ship detection using satellite enhance the detection of vessels which often don't need any tracking system installed on it like fishing boats and majorly the boats which have been in the investigated areas with no permission. Airborne and Satellite-borne data may enable the observation of marine objects remotely, independent of ground circumstances [3]. Target detection and monitoring in the maritime environment are imperative to ensure safety and security on the open sea. One of the most famous maritime surveillance services is the automatic identification system (AIS) for the precise positioning of moving ships [4]. However, most of the vessels and small boats are not equipped with AIS transceivers. Moreover, some vessels turn off their receivers to execute illegal activities, making it harder to detect them. This paper presents the vessel detection from the SAR data and Convolution Neural Network (CNN) algorithm for better computational analysis [5]. CNN is a deep learning algorithm which finds different patterns within an image with several feature extraction algorithms like Canny Edge Detection, K-Means Clustering on colour pixels, Grey Level Co-occurrence, etc., and then the most contrasting extracted features are filtered with the max-pooling method and then these features are passed down to a chain of multi-layer perceptron to train the model to detect similar patterns when exposed to a different image [6]. Now when the CNN model finds similar patterns over an image, it is then converted into a max-pooling layer which is then matched with the max-pooling layer from the training data and ultimately the detection results are localized to a specific region based on Single Shot Multibox Detector (SSD) algorithm [7]. Constant False Alarm Rate (CFAR) algorithm is a threshold-based detection approach for vessel detection with SAR data and widely used around the world [8]. However, as the algorithm is threshold-based therefore several objects are misclassified with different threshold values. CFAR is prone to both false negative and false positive error, this is because, in case of higher threshold values, ships with lower reflectance are left out where are other objects with high reflectance to SAR are often labelled as ship even though it might be noise. Therefore our proposed technique can solve the issue by introducing smart decision-making system and improve the overall detection reliability.

## 2. DATA AND METHODS

### 2.1. Data

In this study, we utilize the Copernicus Sentinel-1 SAR data for sea observation. EADS Astrium GmbH of Germany has structured and created the C-SAR instrument. The instrument is capable of capturing data throughout the day and night on varying landscape conditions [9]. Sentinel-1 data were acquired over the two unique locations of Indian water bodies, including coasts and ports and both locations were considered for the experiment. There are primarily four methods of obtaining Sentinel-1 data: Interferometric Wide (IW), Extra Wide (EW), and Stripmap (SM). Both vertical (V) and horizontal(H) polarization channels can be transmitted by the radar, and either in H or V mode. Hence, a Sentinel-1 procurement is the subset of the accompanying set mixes: single - HH, VV, HV or VH and double – HV+HH or VH+VV. The experiment has been conducted with IW SAR images with VV+VH polarization. A wavelet transformation has been applied to the SAR data to denoise it before passing it toward the CNN model for detection.

The information selected for executing the study comprises the coastline of India. The Sentinel-1 images are of the Mumbai Colaba port to Navi Mumbai port area, situated on the Arabian Sea denoting the west coast, India (Figure 1). The depictions of the Sentinel-1 satellite are gained on 10$^{th}$ January 2020.

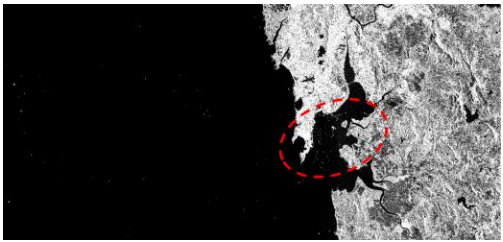

Figure 1: Study location of Mumbai Colaba port to Navi Mumbai port area, INDIA, situated on the Arabian Sea (Source: Copernicus Sentinel data 2020, processed by ESA).

### 2.1. Methods

A Convolution Neural Network (CNN) is a type of deep learning which works on the principle of extraction of contrasting patterns from an image and magnifying the feature to make consistent detection within an image based on similar colour, texture and shapes of objects within the image [10]. In this network several convolution layers are combined with several densely connected layers of neural nodes and the network is trained to identify different patterns from the images. We can depict this graphically as shown in fig. 2.

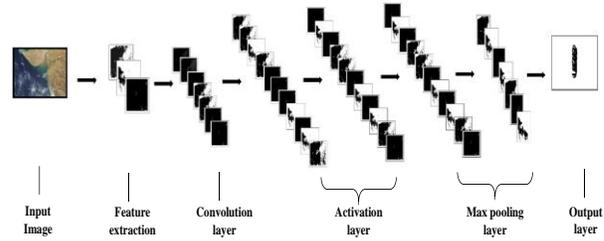

Figure 2: Prepared Convolutional Neural network structure.

The input image is first scanned to find several features and augmentations and then passed through multiple convolution layers to identify several patterns from the image. Chain of activation layers are applied to transform the vectors and most significant features are filtered out with max-pooling layer

We have used 1000 vessel data from SAR images and randomly split them into two parts, 750 for training and 250 for validation of the model. To train the model fast and with little training data and around 200 epochs, we have used SSD MobileNet V2 (COCO) weights and biases to initialize the model. SSD MobileNet V2 (COCO) has been trained on over 300,000 and using the weights and biases of this model to our network, a very little amount of data and epochs are required to converge toward the global minima of loss.

## 3. RESULTS AND DICUSSION

Sentinel-1 acquisitions over the Indian coast visible regions and their ports are considered for the investigation of vessel detection utilizing examination for unidentified maritime objects. Figures 3 and 4 present an example of the detected ships/vessels at Mumbai Colaba port to Navi Mumbai port areas with our proposed algorithm. It is noticed that number of objects distinguished in the VV polarization (red circle) either indicating little ships or vessels.

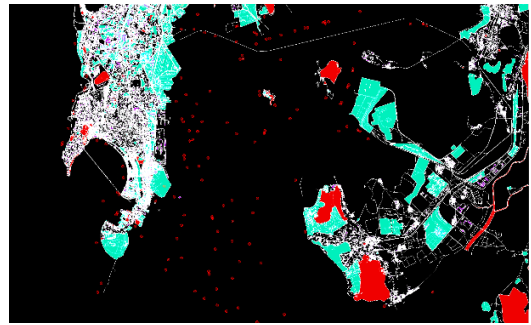

Figure 3: Distribution pattern of the detected ships/vessels at Mumbai Colaba port to Navi Mumbai port area.

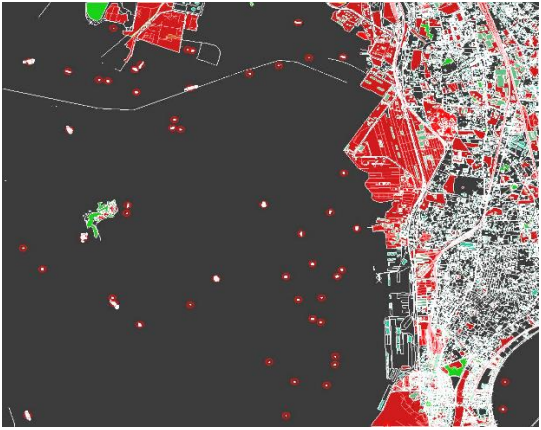

Figure 4: Distribution pattern of the detected ships/vessels at the Navi Mumbai port, India. The red coloured spot in the black background (ocean) depicts marine objects in the coastal area.

Table 1: Detection Performance Summary

|  | Performance |
|---|---|
| Accuracy (%) | 95.46593757120073 |
| Precision | 0.9292220761344402 |
| Recall | 0.948956901990269 |
| F1 Score | 0.9385346764142002 |
| Cohen Kappa Score | 0.8771129475932858 |
| Jaccard Score | 0.8858466739683826 |
| Training Time (ms) | 12953.688621520996 |
| Detection Time (ms) | 3.247499465942383 |

By the proposed approach, we have conducted the training process with 750 vessels data and tested on all 250 vessels data in test images and the observations are recorded in table 1. We have obtained up to 95.46% detection accuracy. We have a higher recall value than precision indicating a lower chance of missing any vessel. The training took 12953.688 milliseconds to converge into a global minimum and 3.247 milliseconds to detect the vessels and this is a property of deep learning that it takes lots of time to update the weights and biases of the nodes to reduce cost ultimately reaching a global minimum and while generating results, as this is simply a matter of simple arithmetic calculations, detection with deep learning is very fast. From these, we can conclude that although deep learning requires plenty of time to train the model, it is still a preferred choice of detection algorithm as the detection is highly accurate and fast and also the model is trained very fast by initializing our model with SSD MobileNet V2 (COCO) weights and biases.

## 4. CONCLUSION

Integrated SAR and CNN-based approaches have provided an improved vessel detection with better clarity and precision. Sentinel-1 is a satellite with promising remote sensing capabilities and its SAR-C band could be utilized for automatized detection of marine objects. CFAR is a popular algorithm associated with SAR to detect marine objects however as the algorithm is threshold-based, it fails to recognize objects out of threshold limits. To overcome this, we have proposed a deep learning CNN model which can be used to smartly detect different marine vessels from SAR images with up to 95.46% accuracy within 3.2 milliseconds timeframe.